\documentclass{article}
\usepackage{microtype}
\usepackage{graphicx}
\usepackage{subfigure}
\usepackage{booktabs}
\usepackage{hyperref}

\usepackage[accepted]{starlab2025}
\usepackage{amsmath}
\usepackage{amssymb}
\usepackage{mathtools}
\usepackage{amsthm}
\usepackage[capitalize,noabbrev]{cleveref}
\theoremstyle{plain}

\theoremstyle{definition}

\theoremstyle{remark}

\usepackage[textsize=tiny]{todonotes}

\starlabtitlerunning{ML For Hardware Design Interpretability: Challenges and Opportunities}

\begin{document}
\twocolumn[
\starlabtitle{ML For Hardware Design Interpretability: Challenges and Opportunities}

\starlabsetsymbol{equal}{*}

\begin{starlabauthorlist}
\starlabauthor{Raymond Baartmans}{sch}
\starlabauthor{Andrew Ensinger}{sch}
\starlabauthor{Victor Agostinelli}{sch}
\starlabauthor{Lizhong Chen}{sch}
\end{starlabauthorlist}
\starlabaffiliation{sch}{School of Electrical Engineering and Computer Science, Oregon State University, Corvallis, Oregon, USA. Contact: \{baartmar, ensingea, agostinv, chenliz\}@oregonstate.edu}
\starlabkeywords{Machine Learning, Large language model, hardware design, EDA, interpretability}
\vskip 0.3in
]

\printAffiliationsAndNotice{} 

\begin{abstract}
The increasing size and complexity of machine learning (ML) models have driven the growing need for custom hardware accelerators capable of efficiently supporting ML workloads.
However, the design of such accelerators remains a time-consuming process, heavily relying on engineers to manually ensure design interpretability through clear documentation and effective communication.
Recent advances in large language models (LLMs) offer a promising opportunity to automate these design interpretability tasks, particularly the generation of natural language descriptions for register-transfer level (RTL) code, what we refer to as ``RTL-to-NL tasks.''
In this paper, we examine how design interpretability, particularly in RTL-to-NL tasks, influences the efficiency of the hardware design process. We review existing work adapting LLMs for these tasks, highlight key challenges that remain unaddressed, including those related to data, computation, and model development, and identify opportunities to address them.
By doing so, we aim to guide future research in leveraging ML to automate RTL-to-NL tasks and improve hardware design interpretability, thereby accelerating the hardware design process and meeting the increasing demand for custom hardware accelerators in machine learning and beyond.
\end{abstract}

\section{Introduction}

\begin{figure}
    \centering
    \includegraphics[width=1\linewidth]{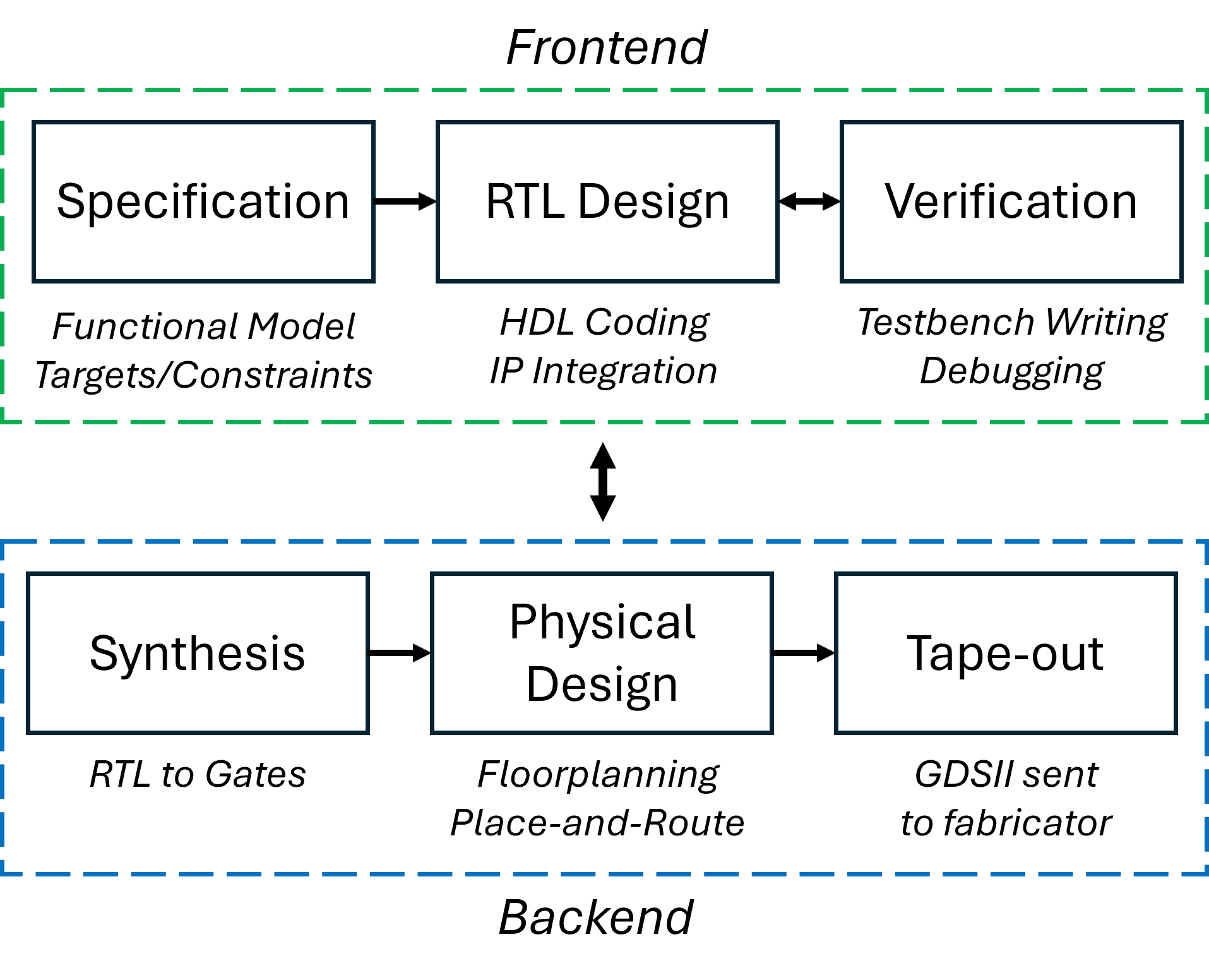}
    \caption{The standard VLSI design flow consists of a front-end phase, where functionality is specified, implemented in RTL, and verified, followed by a back-end phase, where the design undergoes synthesis, physical design, and final preparation for fabrication.}
    \label{fig:vlsiflow}
\end{figure}

A significant area of research and industry attention has focused on the development of custom hardware accelerators to address critical challenges in the computational efficiency and scalability of machine learning (ML) applications.
The increasing size and complexity of modern deep learning models have led to significantly higher computational demands for training and inference, a challenge that has become especially pronounced with the rise of large language models (LLMs) in recent years \citep{radford_2019_gpt2, brown_2020_gpt3, openai_2024_gpt4}. For example, training OpenAI's GPT-3 \cite{brown_2020_gpt3} required approximately $3.14 \times 10^{23}$ floating-point operations (FLOPs), highlighting the immense computational resources needed for such large-scale models. 
While approaches such as model quantization \cite{fan_2020_quant_noise, dettmers_2022_int8, tseng_2024_quip} and algorithmic optimization \cite{ainslie_2023_gqa, yang_2024_gated, agostinelli_2024_leapformer, deepseekai_2025_deepseekr1} can drastically reduce resource consumption, just how much optimization can be achieved is limited by what the underlying hardware is designed to support. 
One such example is BitNet \cite{wang_2023_1bit, ma_2024_bitnet, wang_2024_a4.8}, a ternary weight LLM which, although matching the performance to various full-precision LLMs, has limited efficiency gains since ternary arithmetic is not well supported on general-purpose GPU hardware. 
In contrast, specialized hardware accelerators can enable such models to run more efficiently by eliminating unnecessary components from the chip, freeing up chip area to implement logic specifically designed for specialized functions like ternary arithmetic.
This has created a growing demand for research and development of custom ML hardware \cite{eyeriss_2017, reuther2020}, designed to efficiently handle the unique requirements of modern machine learning workloads.

The increasing demand for custom hardware accelerators has also driven the need for more efficient workflows in hardware development.
An overview of the hardware design cycle, often referred to as the VLSI flow, is shown in Figure \ref{fig:vlsiflow}.
During the ``frontend'' phase, a design specification and its various sub-components must be expressed in register-transfer level (RTL) logic using a hardware description language (HDL) such as Verilog. 
While electronic design automation (EDA) tools now exist to automate many rote or repetitive aspects of HDL coding, the majority of effort in design cycles is spent on integrating sub-components and verifying the correctness of the overall design \cite{chen_2017_soc_trends, varambally2020}.
These processes rely on \textbf{design interpretability}, where each sub-component must be clearly specified and communicated to designers, who must then ensure their RTL implementations are well-documented and accurately shared with other teams to ensure proper integration and functionality.
Given the increasing complexity and scale of modern hardware designs \cite{Ernst01092005, Ziegler2017, ponte_vecchio_intel}, providing clear natural language descriptions for every part of a design is a slow and challenging task, especially for globally distributed engineering teams with many different linguistic backgrounds.
In this paper, we primarily focus on \textbf{RTL-to-NL tasks} as a key aspect of design interpretability, where machine learning models are applied to generate natural language explanations of RTL code.

Applying machine learning-based natural language processing (NLP) methods for design interpretability could lead to significant speed-ups across the hardware design cycle. 
Recently, large language models (LLMs) have demonstrated effective performance at both natural language \cite{hendrycks2021mmlu, xiong2023, saba2024, zhang-etal-2024-benchmarking} and coding-related tasks \cite{tarassow2023, nijkamp2023, bairi2024, chen2024}, including code documentation \cite{luo2024repoagent, nam2024using}.
As interest grows in adapting LLMs for hardware design tasks \cite{zhang2024mgverilog, thakur_vgen,  zakharov2024hdleval}, they appear poised to become a critical tool for design interpretability and the automation of RTL-to-NL tasks.

In this paper, we explore the challenges and opportunities of applying machine learning to hardware design interpretability and RTL-to-NL tasks.
We highlight the critical importance of design interpretability and RTL-to-NL tasks in chip design cycles (Section \ref{sec:design_interpretability_rtl_to_nl}).
We survey the limited existing research in this area and other related topics (Section \ref{sec:current_work}).
We discuss outstanding dataset, computational, and model development challenges that remain unaddressed by existing work (Section \ref{sec:challenges}), and identify opportunities for addressing these challenges (Section \ref{sec:opportunities}).
Finally, we include further discussion the long-term significance of this emerging area of research (Section \ref{sec:further_discussion}).
In doing so, we hope to guide future research in using ML to automate RTL-to-NL tasks and hardware design interpretability, which in turn can greatly accelerate hardware design processes.

\section{Design Interpretability and RTL-to-NL Tasks} \label{sec:design_interpretability_rtl_to_nl} 

Design interpretability (not to be confused with model interpretability) refers to the human-readability of hardware designs, ensuring that the RTL or other design representations can be easily understood, communicated, and modified by engineers.
In this section, we introduce the RTL-to-NL task, and highlight the importance of design interpretability across the design cycle.
Afterwards, we provide a brief outline of the infrastructure needed to apply machine learning to this problem.

\subsection{RTL-to-NL Tasks}
\label{sec:rtl2nl}
Within the broader umbrella of design interpretability, we define RTL-to-NL tasks as tasks which require interpreting RTL designs and related artifacts (e.g., simulator logs, error messages) to produce a response in natural language. 
RTL designs are expressed using hardware description languages (HDLs) such as Verilog, which differ from traditional programming languages in that they model the structure of a digital circuit, including critical aspects like concurrency and timing.

While it may seem initially unclear how such concepts could be accurately communicated in natural language, natural language is actually well-equipped for describing RTL. Describing timing information or concurrency in natural language can be accomplished by properly using phrases such as “at the same time,” “meanwhile,” or “until.” The fact that engineers routinely explain and teach RTL concepts in textbooks, documentation, and discussions using natural language underscores this. 

\begin{figure}[h]
    \centering
    \includegraphics[width=1\linewidth]{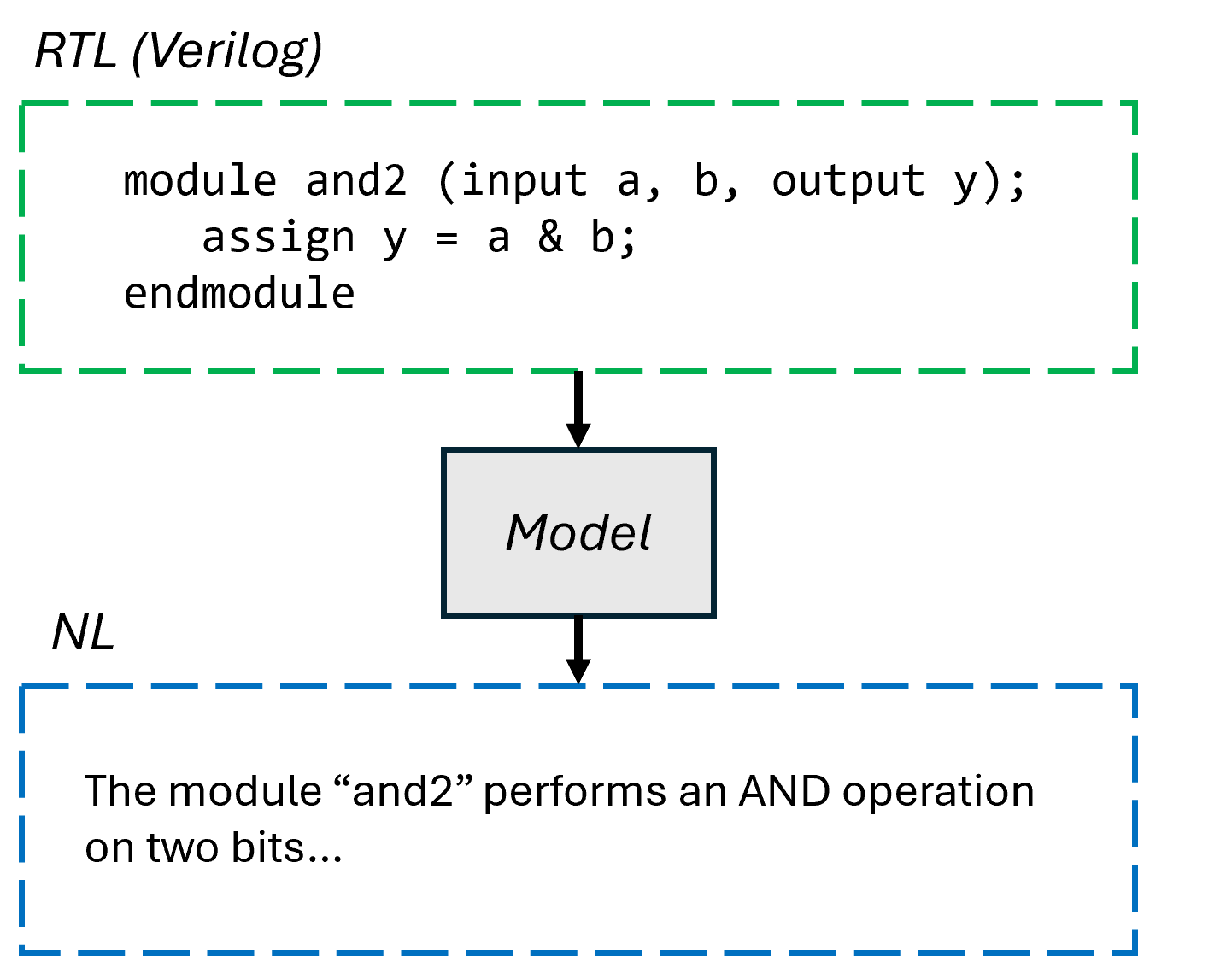}
    \caption{An example of a possible RTL-to-NL task, where the goal is to generate natural language documentation for an RTL design described in Verilog.}
    \label{fig:rtl2nl}
\end{figure}

Simple examples of RTL-to-NL tasks would be writing documentation for an undocumented RTL design, or answering a high-level question like ``\textit{How does this design work?}''
A more complex version of an RTL-to-NL task might involve reasoning about specific behaviors within the design, such as answering, ``\textit{What control signals does this design use to ensure that data dependencies are resolved in the matrix multiplication unit?}''
Figure \ref{fig:rtl2nl} illustrates an example of an RTL-to-NL task, where a model (e.g., an LLM) generates a specification based on a given Verilog design.
Unlike traditional code-to-NL tasks, which describe a sequence of instructions to be executed on a CPU/GPU, RTL-to-NL tasks are unique in multiple ways:
\begin{enumerate}
    \item \textbf{RTL represents hardware structure and behavior.} While typical software code defines sequential algorithms, RTL (e.g., Verilog) directly determines the structure and behavior of a digital circuit. For example, generating a natural language description of a Verilog counter requires not only explaining its behavior (incrementing a counter) but also describing the underlying hardware, such as the 4-bit register and adder that operate synchronously on each clock cycle.
    \item \textbf{Distinct syntactic structure of RTL.} For example, while state-updating operations in software are typically expressed as sequential function calls or object state modifications, in Verilog, logic is typically defined within always blocks. These blocks execute concurrently, independent of each other, and are triggered by specific sensitivity conditions, such as clock edges or signal changes. Syntactic structures like this must be handled separately than typical software code.
    \item \textbf{Understanding RTL requires foundational knowledge in digital logic.} Reading RTL requires a fundamental understanding of hardware design and digital logic concepts. As a result, LLMs for RTL-to-NL tasks must also possess foundational knowledge of these concepts. Figure \ref{fig:simple_example} illustrates how typical LLMs, which are primarily trained on software code, struggle to answer basic questions that require an understanding of digital logic design.
\end{enumerate}

\subsection{Impact of Interpretability in the Design Cycle}
\label{sec:need_interpretability}
As hardware designs continue to grow in scale and modularity \cite{Ernst01092005, Ziegler2017, ponte_vecchio_intel}, understanding the vast range of interconnected subsystems has become increasingly challenging.
In this context, RTL interpretability can have a significant impact on the speed of the design cycle.

\subsubsection{Legacy Code and IP Integration}
\label{sec:ip_integration}

Understanding code written by others is often cited as one of the most challenging and time-consuming aspects of engineering projects \cite{sun2023surveysourcecodesearch, 10670481}, and chip design is no exception. Modern chip designs now often consist of many smaller subcomponents (IP) developed by individual teams or external vendors \cite{Kim2020, ipxact_ieee}. When integrating IP with missing or poor documentation, engineers may have to reverse-engineer the module's functionality and behavior, significantly slowing down the integration process. In contrast, having well-documented IP can greatly accelerate the integration process, as evidenced by the industry-wide adoption of standardized IP description formats \cite{ipxact_ieee}.

\subsubsection{Verification}
\label{sec:verification}

Verification, in which a design is tested for correctness, is the most time-consuming part of hardware development, occupying up to 70\% of the entire chip design cycle \cite{varambally2020}.
Surveys have found that debugging alone can take up 44\% of this process \cite{Mutschler_2021}, meaning that up to a third of the entire design cycle may be spent on RTL-to-NL tasks for debugging, such as reading RTL code, simulator logs, and other design artifacts for fault localization.
Importantly, time spent in debug more than doubles the time spent on any other verification task, such as running simulations (21\%), testbench development (19\%), and test planning (13\%). Misunderstandings or lack of clarity in RTL documentation can further extend the time spent on debugging. The quicker engineers can access accurate RTL-to-NL translations, the more time they can spend resolving actual issues.

\subsubsection{Electronic Design Automation}
\label{sec:EDA}

\begin{figure}
    \centering
    \includegraphics[width=1.0\linewidth]{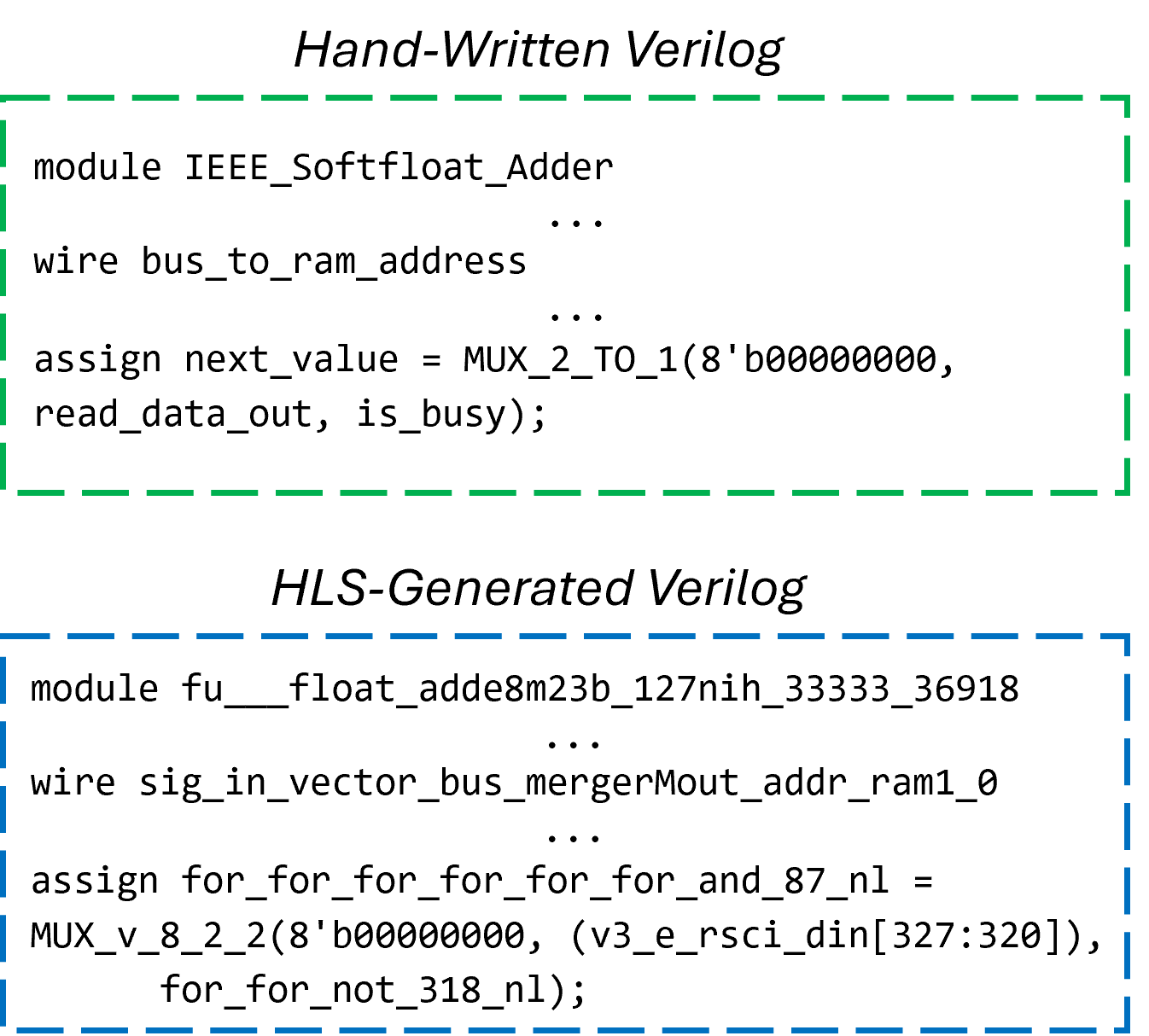}
    \caption{An example of how hand-written Verilog may compare to the code generated by High-Level Synthesis tools. The HLS-generated code in the figure was generated by Bambu HLS \cite{bambu} as well as Catapult HLS \cite{catapult_hls}. }
    \label{fig:RTLhandvsHLS}
\end{figure}

Electronic Design Automation (EDA) tool developers have long sought to automate as many chip design tasks as possible.
However, this pursuit of automation can come at the expense of human readability and interpretability of the tool's outputs.
High-Level Synthesis (HLS) is one such tool, offering the potential to significantly reduce the need for manual RTL coding by automatically translating a functional specification written in C into RTL.
However, these HLS tools prioritize performance over readability and maintainability, as shown in Figure \ref{fig:RTLhandvsHLS}. 
Variable names are often incredibly long and unintelligible, the design logic is hard to follow, and the tools do not optimize the organization of the code or design hierarchy for better maintainability. 
Although this can cause challenges when designing for field-programmable gate arrays (FPGAs), it becomes an even greater problem when designing for large-scale fixed application-specific integrated circuits (ASICs). 
ASIC design flows, where the majority of designers primarily work at the RTL level, often need to meet much stricter performance, power, or area targets.
Since HLS tools do not guarantee optimal solutions, their generated RTL is less likely to be used in these flows, even as an initial RTL draft, because its lack of readability and RTL documentation makes any further hand-tuning nearly impossible.
RTL-to-NL models could generate the necessary documentation and explanations to make the outputs of automated design tools more understandable, thereby expanding the potential applications of these tools to accelerate design cycles.

\subsubsection{AI Alignment in Chip Design}
\label{sec:human_ai_collab}
With the rise of generative AI tools powered by large deep learning models, the need for design interpretability has become increasingly important.
As these AI-driven tools become integrated into chip design workflows, maintaining human interpretability of their outputs is essential to ensure that generated designs align with engineers' expectations.
This is particularly crucial for preventing scenarios where an AI might introduce vulnerabilities into a design.

\subsection{Needed Infrastructure}
\label{sec:needed_infrastructure}
These challenges highlight the importance of developing tools which can provide automated support for RTL-to-NL tasks and design interpretability.
Traditional approaches, such as rule-based systems and static analysis tools \cite{Bajwa2016, chiticariu-etal-2013-rule, XU2021101288}, rely on extensive manual rule engineering and often struggle to generalize beyond narrowly defined domains. In contrast, deep learning models—particularly large language models (LLMs)—have demonstrated strong generalization capabilities \cite{kang2024, otherwang2024, liu2024, openai_2024_gpt4} and have outperformed rule-based methods in various code analysis tasks \cite{LU2024112031, 10706805}.
Reflecting this potential, NVIDIA’s internal studies on hardware design workflows have identified LLM-based assistants as a promising way to boost engineering productivity \cite{liu_2024_chipnemo}.
At the same time, the widespread adoption of standardized IP formats such as IP-Xact \cite{ipxact_ieee} signals a broader industry consensus on the importance of improving design interpretability to support more efficient and scalable hardware development cycles.

However, adapting models for RTL-to-NL tasks requires dedicated ML infrastructure to support training and evaluating models on RTL-to-NL tasks, including appropriate datasets, evaluation benchmarks, fine-tuning frameworks, and more. Despite growing interest in applying LLMs to hardware design, this supporting infrastructure remains underdeveloped. In the next section, we examine the existing efforts to address these gaps.

\section{Current Work} \label{sec:current_work}

In this section, we survey the existing research on adapting LLMs for hardware design interpretability and RTL-to-NL tasks.
We also briefly summarize relevant work on related topics.

\subsection{LLM for RTL-to-NL Tasks}
\label{sec:current_work_rtlnl}

\subsubsection{DeepRTL}
\label{sec:deeprtl}
\textit{DeepRTL} \cite{liu2025deeprtl} presents the first dataset and benchmark for assessing LLM performance in both Verilog understanding and Verilog generation.
Their approach leverages GPT-4, along with professional hardware engineers, to annotate a dataset of Verilog designs with natural language descriptions of their functionality.
The annotation takes place at multiple levels of granularity; starting from line-level commenting and eventually reaching high-level block and module-level descriptions.
To fine-tune LLMs on their dataset, a curriculum learning approach is employed, which organizes training into sequential phases for line-level, block-level, and module-level annotations.
To evaluate their dataset, they construct a benchmark of 100 Verilog designs with expert-written descriptions, measuring the embedding similarity of model-generated descriptions against the human references.
Additionally, they measure a “GPT Score,” utilizing GPT-4 to evaluate the quality of the generated description by providing a score between 0 and 1, based on the similarity to the reference description. 

\subsubsection{SpecLLM}
\label{sec:specllm}
\textit{SpecLLM} \cite{li2024} compiles a dataset of hardware architecture specifications from various open-source hardware projects to fine-tune LLMs for generating specifications of RTL designs. Although the dataset can be very useful in many scenarios, one drawback is that it primarily consists of architecture-level specifications, which do not necessarily capture RTL-level details. 
As a result, the improvement in the model's understanding of RTL through this dataset may be limited, as it appears to focus more on architecture standards and specification formatting. The paper does not provide any quantitative evaluation to assess the model's performance after fine-tuning.

\subsubsection{ChipNeMo}
\label{sec:chipnemo}
NVIDIA's \textit{ChipNeMo} \cite{liu_2024_chipnemo} is a general-purpose hardware design assistant LLM, trained on a diverse set of proprietary chip-design documents, including RTL code, specifications, and documentation.
The model incorporates techniques like domain-specific tokenization and retrieval-augmented generation (RAG) to enhance its efficiency.
While the paper does not specifically evaluate its performance on RTL-to-NL tasks, its broad (albeit proprietary) training dataset and general-purpose design make it relevant for inclusion here.
Furthermore, the evaluation/benchmark details are limited, with much of the assessment relying on expert grading.

\subsection{LLM for NL-to-RTL Tasks}
\label{sec:current_work_nlrtl}
While the adaptation of LLMs for RTL-to-NL tasks and hardware design interpretability remains underexplored, adapting LLMs for ``NL-to-RTL'' tasks, such as Verilog generation, has been the dominant focus of prior work \cite{liu2023verilogevalevaluatinglargelanguage, lu2023rtllmopensourcebenchmarkdesign, allam2024rtlrepo, SADS2024dataset, zhang2024mgverilog, thakur_vgen,  zakharov2024hdleval, sandal2024zeroshotrtlcodegeneration, delorenzo2024makecountllmbasedhighquality, delorenzo2024creativevalevaluatingcreativityllmbased}.
A number of works have released datasets and benchmarks to assess LLM-generated Verilog designs, including RTLCoder \cite{liu2024rtlcoder}, VerilogEval \cite{liu2023verilogevalevaluatinglargelanguage} (benchmark only), RTLLM \cite{lu2023rtllmopensourcebenchmarkdesign}, and more \cite{thakur_vgen, chang2024NLRTL, zhang2024mgverilog}.
These works primarily draw upon open-source RTL designs from platforms like GitHub and HDLBits \cite{hdlbits} to construct their datasets.
Other areas of exploration include generating SystemVerilog verification constructs \cite{sun2023towards}, designing in HLS \cite{huang2024new}, EDA tool script generation \cite{liu_2024_chipnemo}, and applying agent-based approaches to hardware design \cite{huang2024towards}.

\subsection{LLM for Coding}
\label{sec:current_work_code_llms}
Outside of hardware design, a significant amount of work has focused on adapting LLMs for coding in more widely used programming languages. 
While it is beyond the scope of this paper to examine all approaches, one relevant technique involves custom attention mechanisms that incorporate abstract syntax trees (ASTs) and language-specific information \cite{ahmad-etal-2020-transformer, wang-etal-2021-codet5, graphcodebert}, which could similarly be extended for LLMs adapted for Verilog processing.
Additionally, we discuss later in Section \ref{sec:learning_challenges} why approaches to adapting LLMs for software coding may not necessarily translate to the realm of hardware design.

\section{Challenges}
\label{sec:challenges}

In this section, we examine outstanding challenges in adapting LLMs for RTL-to-NL tasks which have yet to be fully addressed by existing work.

\begin{table}[h!]
\label{tab:file_counts}
\centering
\begin{tabular}{l r}
\toprule
\textbf{Language} & \textbf{Number of Files} \\
\midrule
Java & 155,000,000 \\
C++  & 104,000,000 \\
Python & 95,400,000 \\
Go & 24,100,000 \\
\textbf{Verilog} & \textbf{1,200,000} \\
\textbf{SystemVerilog} & \textbf{465,000} \\
\bottomrule
\end{tabular}
\caption{The number of files on GitHub categorized by programming language, highlighting the disparity between HDLs and popular software languages such as Java and Python.}
\end{table}

\subsection{Dataset Construction}
\label{sec:lack_datasets}
The most critical gap limiting research on adapting LLMs to RTL-to-NL tasks has been the absence of datasets of Verilog code with corresponding natural language labels, such as design specifications or question-answer pairs. Assembling such datasets is challenging due to the competitive nature of the industry and the smaller number of chip designers compared to software engineers, leading to a lack of open-source RTL designs to use for dataset construction. As shown in Table \ref{tab:file_counts}, the total number of Verilog and SystemVerilog files on GitHub is approximately 93 times smaller than the number of Java files. Additionally, the time required to manually label data and the expertise needed for proper labeling further prevent the use of data annotation platforms \cite{Wang2024}, which could otherwise expedite the creation of these datasets. 

While some progress has been made in automating RTL dataset creation by using publically available RTL designs from GitHub \cite{thakur_vgen}, many of the sources that provide reliably structured or labeled data, such as HDLBits \citep{hdlbits}, mainly feature basic problems that do not reflect a broader set of real-world designs. As such, recent works on NL-to-RTL \cite{liu2024rtlcoder} and RTL-to-NL \cite{liu2025deeprtl} tasks tend to heavily rely on synthetic data to compensate for the lack of high-quality, labeled examples.

\subsection{Evaluating RTL-to-NL Accuracy}
\label{sec:evaluating_accuracy}
While NL-to-RTL tasks can be evaluated by running LLM-generated code against human-written test suites to verify functional correctness, evaluating RTL-to-NL tasks is less straightforward. Natural language descriptions or specifications of hardware must convey critical implementation details, where even minor omissions or inaccuracies can significantly distort meaning.
Even when data is available, research on RTL-to-NL tasks still lacks robust approaches for accurately evaluating the quality of natural language documentation. 
Existing reference-based measures such as BLEU or ROUGE, which target lexical similarity, could give misleadingly high scores to critically different descriptions simply due to lexical overlap. Similarly, embedding similarity can compare if an LLM-generated description is semantically close (i.e., uses many semantically related words, like “clock,” “reset,” etc.), but cannot penalize omissions or vague descriptions that could lead to serious implementation mismatches.
An alternative way to evaluate accuracy could be through expert grading, but this approach is difficult to scale due to the specialized knowledge required and the inherent subjectivity of graders. Methods like 'GPT-scoring,' such as the one used by \cite{liu2025deeprtl}, are also problematic, as pointwise scoring with LLMs is challenging to calibrate \cite{desai2020calibrationpretrainedtransformers} and tends to be computationally expensive.

\subsection{Computational Challenges}
\label{sec:computational_challenges}
Adapting LLMs to process RTL designs also presents a computational challenge, as representing hardware logic in RTL often requires substantially more lines of code—and consequently, greater context length—compared to describing equivalent functionality in high-level languages like C.
Table \ref{tab:token_comparison} shows that while algorithms are relatively simple to describe in C, their RTL implementations require significantly more code. To make this comparison, we retrieved corresponding C and Verilog implementations of some well-known algorithms from GitHub.
Additionally, we ran the C implementations through Bambu HLS \cite{bambu} to generate RTL, which is shown in the "Verilog (HLS)" column for comparison.
We observe that hand-written Verilog implementations can require more than triple the number of tokens compared to their C counterparts.
Additionally, the length of a single IP generated by HLS can exceed the limits of even commercially deployed models that are optimized for long contexts, such as LLaMa-3 \cite{grattafiori2024llama3herdmodels} (up to 128,000 tokens), GPT-4 \cite{openai_2024_gpt4} (up to 128,000 tokens), and Claude 3 \cite{claude_3_report} (up to 200,000 tokens).
These context limitations make it clear that scaling to larger designs, which incorporate many different IPs, can quickly become computationally prohibitive.

\begin{table}[h]
    \centering
    \begin{tabular}{lrrr}
        \toprule
        \textbf{Algorithm} & \multicolumn{3}{c}{\textbf{Number of Tokens}} \\
        \cmidrule(lr){2-4}
         & \textbf{C} & \textbf{Verilog} & \textbf{Verilog (HLS)} \\
        \midrule
        GEMM & 252 & 18423 & 224289 \\
        Sobel & 557 & 2031 & 39849 \\
        BellmanFord & 1440 & 5156 & 47335 \\
        SHA256 & 2683 & 9290 & 260356 \\
        \bottomrule
    \end{tabular}
    \caption{The token counts for implementing a variety of algorithms in C and Verilog. The "Verilog (HLS)" column represents Verilog code generated by Bambu HLS from the corresponding C program. Token counts were obtained using the GPT-4 tokenizer \citep{openai_2024_gpt4}.}
    \label{tab:token_comparison}
\end{table}

Moreover, LLMs are known to suffer performance degradation when handling long context lengths \cite{anonymous2024longcontext, an2024doeseffectivecontextlength, laban-etal-2024-summary}, and exhibit a tendency to lose focus on middle-context information \cite{liu-etal-2024-lost}.
They also demonstrate a tendency to ``hallucinate'', generating outputs that may appear coherent but are nonsensical or inaccurate \cite{ji-etal-2023-towards,friel2023chainpollhighefficacymethod}.
The effects of these properties can be particularly detrimental for interpreting RTL, where tracking the code's structure and dependencies is essential for correctly understanding the design and ensuring accurate interpretation of its functionality. 
The current performance and reliability challenges of LLMs, particularly when dealing with long contexts, make research in adapting LLMs for RTL-to-NL tasks more challenging, as it would be difficult to demonstrate their applicability to real-world RTL development, where designs may span hundreds of thousands of tokens.
Critically, many of the currently existing benchmarks for both Verilog generation and Verilog understanding \cite{chang2024NLRTL, liu2025deeprtl} do not include designs exceeding 2048 tokens in length.


\subsection{Training/Learning Challenges}
\label{sec:learning_challenges}
As we previously discuss in section \ref{sec:rtl2nl} HDLs, such as Verilog, are fundamentally different from traditional imperative programming languages, as they are used to describe the structure of digital circuits, rather than defining sequential behavior.
This poses unique challenges for LLMs, including those optimized for coding \cite{rozière2024codellamaopenfoundation}, as these models have been primarily trained on sequential programming languages.
Such a model may misinterpret a design's concurrency, assuming two components execute sequentially in the order they appear in the code.

RTL code also involves underlying digital design concepts that are rarely explicitly documented within the code itself but can be essential for understanding the design’s performance and functionality.
Critical path timing, circuit area, and the effects of propagation delays are crucial to accurate design analysis but are not explicitly represented in Verilog.
In hardware design, different logic components exhibit varying delay characteristics.
For instance, an adder generally has higher delay than a basic ``AND'' gate due to the additional gates involved in carry propagation.
Now consider a scenario where a model trained on traditional software-oriented code is asked to document the critical path in a Verilog design. Without an understanding of component-specific delays, such a model might naively identify the longest dependency chain as the critical path, focusing solely on operation count. In doing so, it could miss a shorter path that includes higher-delay components, such as adders, which ultimately defines the actual critical path.

\begin{figure}[h]
    \centering
    \includegraphics[width=1\linewidth]{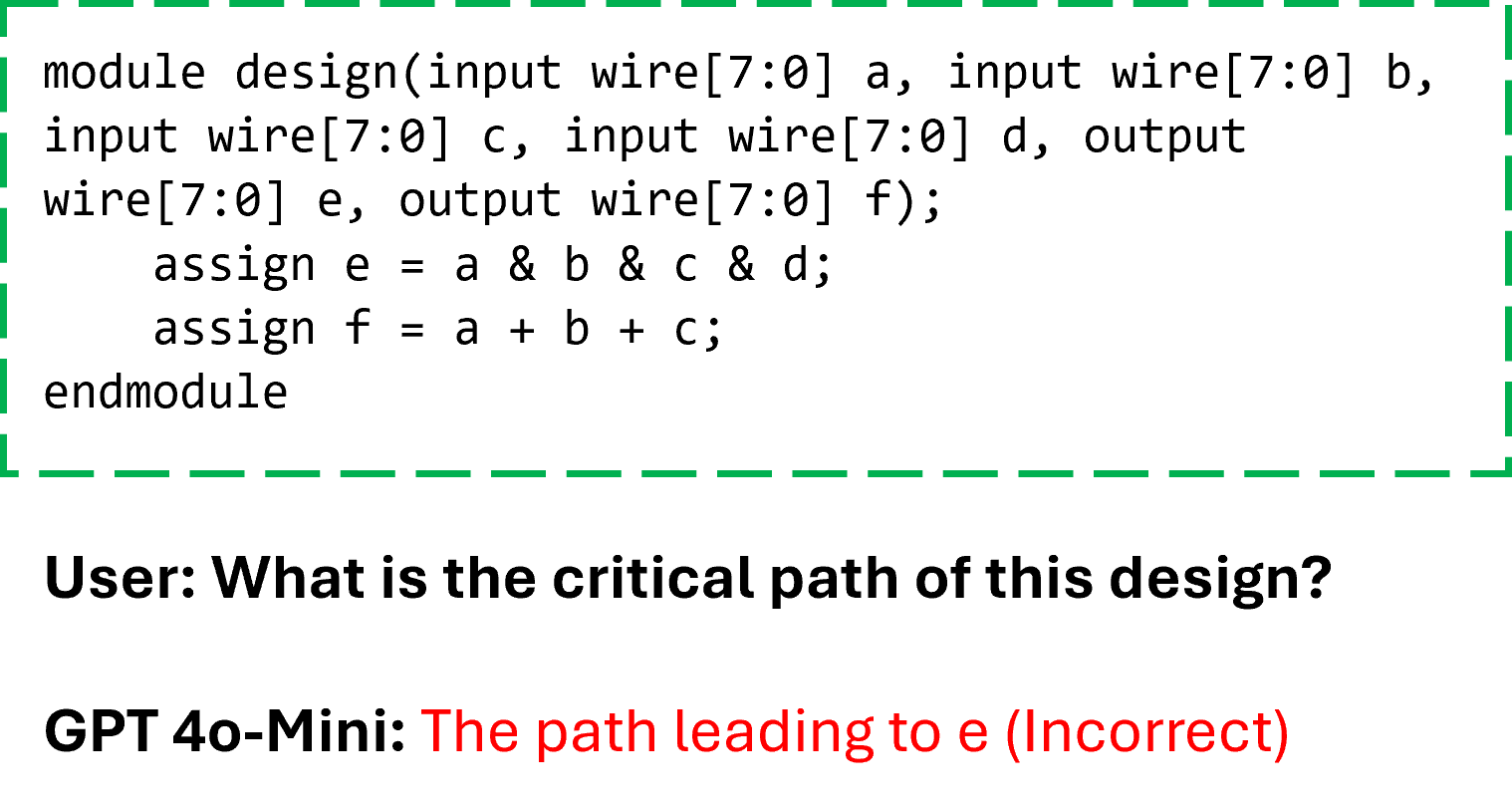}
    \caption{We asked GPT4o-mini to evaluate the critical path of simple Verilog modules. This figure shows how the model will give incorrect or inconsistent answers. For this example, the path leading to f will have a longer critical path due to carry logic necessary for the adders.}
    \label{fig:simple_example}
\end{figure}

Figure \ref{fig:simple_example} shows how GPT 4o-mini can make this exact mistake when asked to identify the critical path in a simple Verilog design.
In the figure, GPT selected the path leading to \textit{e}, likely because it contained more operations than the path leading to \textit{f}, even though the latter has a longer critical path due to the carry logic involved in the adders.
Importantly, RTL code that requires the most interpretation is often poorly organized, featuring inconsistent or poorly named variables that do not align with human-readable conventions.
As a result, adapting LLMs for RTL-to-NL tasks requires training approaches that not only enable the model to learn the unique structure of the HDLs used to describe RTL designs and underlying hardware design concepts, but also enable it to interpret poorly organized RTL code that lacks meaningful semantic clues.

\section{Research Opportunities}
\label{sec:opportunities}
The challenges outlined in the previous section present many opportunities where new research can help provide the critical ML infrastructure outlined in Section \ref{sec:needed_infrastructure}. This infrastructure could catalyze greater research participation towards adapting LLMs for RTL-to-NL tasks, creating new tools that can drastically improve the efficiency of hardware development.

\subsection{Approaches for Dataset Construction}
\label{sec:creating_datasets}
Datasets are the foundational infrastructure needed to adapt LLMs for RTL-to-NL tasks. However, creating task-specific RTL-to-NL datasets is complex and time-consuming, particularly when manual labeling is required, which makes it challenging for most research groups to undertake. Therefore, innovative approaches to automated dataset generation are essential. Recent works on NL-to-RTL \cite{liu2024rtlcoder} and RTL-to-NL \cite{liu2025deeprtl} tasks have started to explore synthetic data as a solution to the lack of high-quality labeled examples. While synthetic data has proven effective in these works, as well as in other domains \cite{math10152733, lu2024machinelearningsyntheticdata, openai_2024_gpt4}, it is not guaranteed to be effective, and further exploration is needed to determine how such methods could be successfully applied for RTL-to-NL tasks. One potential approach could integrate HLS or RTL synthesis tools with LLMs like GPT to automatically generate and label question-answer pairs on critical RTL design aspects, such as concurrency.

Furthermore, there is still significant potential for community-driven efforts to create and expand RTL-to-NL datasets. The newly established IEEE International Conference on LLM-Aided Design \cite{iclad} features a datasets track, which could help bring together researchers interested in advancing RTL-to-NL datasets through collaboration.

\subsection{New Approaches for Evaluation} \label{sec:approaches_evaluation}
As discussed in Section \ref{sec:evaluating_accuracy}, evaluating RTL-to-NL accuracy is a nontrivial problem where existing metrics are likely inadequate. One potential approach to evaluating LLM-generated explanations is to have another LLM regenerate the same RTL design based on the explanation, followed by logical equivalence checking (LEC) to compare the regenerated design with the original. This approach is inspired by the concept of Round-Trip Correctness \cite{allamanis2024unsupervisedevaluationcodellms}, where an LLM generates a natural language description of a subset of code, then re-implements the code in the target language. In the case of RTL-to-NL, a separate LLM with adequate Verilog skills could be used to re-implement the design. This approach could provide a better signal of whether critical design details are accurately captured in natural language, reducing the reliance on human-written "ground truth" references. However, challenges remain, such as distinguishing between hallucinations/coding errors and inaccuracies in the description. Another approach could explore distilling expert grading into a BERT model to approximate expert judgment, similar to the COMET model used in machine translation \cite{rei2020comet}.

\subsection{Model Architectures for Handling Long Contexts of RTL}
\label{sec:modelarchs_longcontext}
Efficiently processing long context lengths is crucial for any LLM dealing with extensive RTL code. As outlined in Section \ref{sec:computational_challenges}, RTL designs often involve large context spans, making it challenging for models to capture dependencies across the entire design hierarchy.
One approach to addressing this challenge is modifying attention mechanisms to take advantage of the hierarchical and modular nature of RTL, similar to techniques used in code-specific transformer models \cite{ahmad-etal-2020-transformer, wang-etal-2021-codet5, graphcodebert}. These models incorporate structural information about the code directly into the attention function. Such modifications could enhance recall over long contexts by focusing on relevant signal dependencies and module interactions, ultimately improving the model’s ability to handle complex RTL designs.

Additionally, RTL code often contains highly repetitive patterns that require less processing effort, such as the statement \texttt{always @(posedge clk) begin}, which appears in every Verilog module involving sequential logic. While it breaks into six tokens using the GPT-4o tokenizer, its semantic contribution is minimal. Alternative tokenization methods, like Byte-Latent Transformers (BLTs) \cite{pagnoni2024bytelatenttransformerpatches}, adaptively allocate compute to more informative parts of the sequence, improving efficiency.
A custom tokenizer tailored to RTL text data could also reduce context length by condensing repetitive constructs into single tokens. For example, \texttt{always @(posedge clk) begin} could be reduced to one token. NVIDIA's ChipNemo \cite{liu_2024_chipnemo} incorporated a custom tokenizer which demonstrated modest improvements in tokenization efficiency (1.6\% to 3.3\%) on their internal datasets, though public data saw little to no gain.

While not strictly a model architecture, domain-specific retrieval-augmented generation (RAG) techniques \cite{liu_2024_chipnemo} could also help alleviate the burden of long context lengths. By indexing RTL submodules and retrieving only relevant portions of a design during inference, RAG approaches could improve scalability and performance, particularly when dealing with large IP libraries or system-on-chip (SoC) designs. Similarly, another strategy could involve using module embeddings, analogous to document embeddings, to capture the functionality of Verilog submodules in compact vector representations. This could reduce the need to include all submodule definitions in the context, allowing the LLM to operate at a higher level of abstraction. Code embeddings, a well-studied topic \cite{chen2019literature}, could be key in enabling this approach.

\subsection{LLM-Aware EDA Tools}
\label{sec:llm_aware_tools}

As discussed in Section \ref{sec:rtl2nl} and shown in Figure \ref{fig:RTLhandvsHLS}, the RTL code generated by EDA tools, such as HLS, often contains bloat and readability issues that can cause problems for both human designers and LLMs. Code generation tools that enhance readability by producing semantically meaningful variable names, structured hierarchies, and more intuitive representations could significantly improve both human understanding and make the code easier for LLMs to process.

Developers of automation tools may also benefit from recognizing how the intermediate data their tools generate—often discarded after fulfilling its primary function—could serve as a rich source of training data for LLMs.
HLS tools primarily generate Verilog but internally represent designs using graph-based abstractions that guide the scheduling of operations, resolution of data dependencies, and optimization of the hardware datapath. These representations support the step-by-step transformation of functional algorithms into hardware implementations. For example, Bambu HLS \cite{bambu} provides a \texttt{--print-dot} option that lets users visualize these graphs, offering insights about the design that are not easily visible in the Verilog output. Traditionally, this intermediate data is discarded after compilation, but it could be repurposed to train LLMs, enabling them to reason through higher-level hardware design challenges. This approach is aligned with Chain-of-Thought reasoning research \cite{wang2023keqingknowledgebasedquestionanswering, chu2024navigateenigmaticlabyrinthsurvey}, where models enhance performance by breaking down complex problems into intermediate steps.

\section{Further Discussion}
\label{sec:further_discussion}

Given the ongoing uncertainty about how rapidly advancing AI systems will affect the workforce and human roles, we believe it is important to discuss the long-term relevance and potential impact of research on RTL-to-NL tasks and hardware design interpretability in this evolving landscape.

\subsection{Would Fully Autonomous Hardware Design Eliminate the Need for Design Interpretability?}
\label{sec:fully_auto_question}
One may argue that if AI and EDA tools eventually surpass human capabilities across all stages of the design cycle, then making these designs human-interpretable would become unnecessary.
However, while advanced EDA tools have made significant strides, the problem sizes they address are so vast \cite{chen_2018_routerless, Noonan_2024, zhong2023llm4edaemergingprogresslarge} that achieving globally optimal designs is intractable with current technology.
Although deep learning-driven EDA tools have demonstrated substantial improvements in many areas \cite{hafen_2013_eda_and_ml, alphachip, goswami_2023_appl_of_ml}, their models still face significant challenges in being able to produce formal guarantees for correct/optimal output, as verifying even simple neural networks is shown to be NP-hard \cite{katz2017reluplex}. 
As a result, human involvement, including RTL-to-NL tasks, will likely remain essential, even if it eventually serves primarily to provide oversight of AI-generated outputs.
Moreover, as discussed in Section \ref{sec:human_ai_collab}, ensuring design interpretability will be crucial for aligning AI with human engineers' intentions.

\subsection{Is Design Interpretability a ``Human Problem?"}
\label{sec:human_problem}
Another question to raise is that the problems addressed by RTL-to-NL tasks may be fundamentally ``human problems'' and are best solved through better organizational and management strategies, rather then creating complex and expensive ML solutions, which may do nothing but exacerbate existing problems.
For example, knowledge silos across teams and poor documentation for IPs are problems that a design group or organization could resolve by enforcing stricter documentation practices.

While organizational and management improvements are certainly important, this overlooks the limitations of such approaches as hardware design projects grow in scale and complexity.
As teams grow in size and individual engineers take on larger and more intricate subsystems, it becomes increasingly difficult for teams to enforce comprehensive documentation and rigorous standards without micromanagement.
Additionally, manually writing high-quality documentation will always require significant man-hours, which can be alleviated through automation with LLMs \cite{dvivedi_comparative_analysis}.
As discussed in Section \ref{sec:rtl2nl}, LLMs offer notable advantages in generalizability and performance compared to methods like rule-based systems, making them a fitting and practical solution for automating design interpretability.



\section{Conclusion}
\label{sec:conclusion}
The growing computational demands of modern ML models have highlighted the need for custom ML hardware.
However, we find that the hardware design cycle remains heavily dependent on engineers manually ensuring design interpretability through extensive documentation and communication, creating a significant bottleneck. 
Leveraging ML to address this challenge presents a promising opportunity.
In this paper, we examine existing research on this topic and identify the computational and learning challenges that remain unresolved.
We then discuss how these challenges open new research avenues that future research in this area can address.
Critically, progress in adapting LLMs for RTL-to-NL tasks and enhancing hardware design interpretability will not only streamline the hardware development process but could also contribute to broader advancements in machine learning techniques with applications across multiple research domains.

\nocite{langley00}

\bibliography{custom}
\bibliographystyle{starlab2025}

\newpage
\onecolumn

\end{document}